\documentclass{article}
\usepackage{spconf,amsmath,epsfig}
\usepackage{longtable}
\usepackage{xurl}

\title{A Latent Space Metric for Enhancing Prediction Confidence in Earth Observation Data}

\name{I. Pitsiorlas$^{1}$, ~A. Tsantalidou$^{2}$, ~G. Arvanitakis$^{2}$, ~M. Kountouris$^{1}$, ~Ch. Kontoes$^{2}$}
\address{\hspace{5em} EURECOM, France$^{1}$ \hspace{5em} National Observatory of Athens, Greece$^{2}$\\
Communication Systems Department \hspace{4em}  BEYOND Center of Excellence}

\begin{document}
%
\maketitle
\begin{abstract}
A new approach for estimating confidence in machine learning model predictions, specifically in regression tasks utilizing Earth observation data with a particular focus on mosquito abundance (MA) estimation, is proposed here. We leverage the Variational AutoEncoder architecture to derive a confidence metric by the latent space representations of Earth observation datasets. This methodology is pivotal in establishing a correlation between the Euclidean distance in latent representations and the absolute error in individual MA predictions. Our study focuses on Earth observation datasets from the Veneto region in Italy and the Upper Rhine Valley in Germany, considering areas significantly affected by mosquito populations. A key finding is a notable correlation of \textbf{0.46} between the absolute error of MA predictions and the proposed confidence metric. This correlation signifies a robust, new metric for quantifying the reliability and enhancing the trustworthiness of the AI/ML model predictions in the context of both Earth observation data analysis and mosquito abundance studies.
\end{abstract}

\begin{keywords}
Trustworthy AI, latent space representation, VAE, confidence estimation, earth observation data, mosquito abundance.
\end{keywords}
\section{Introduction}
\label{sec:intro}
Vector-borne diseases (VBDs), transmitted through the bites of infected vectors such as mosquitoes, contribute to over 17\% of all infectious diseases, leading to upwards of 700.000 deaths each year \cite{whoGlobalVector}. Mosquitoes are well-known for spreading illnesses such as malaria, West Nile Virus (WNV), Chikungunya, dengue, and Zika through their bites. Despite significant control measures for many of these diseases, recent decades have seen a resurgence in mosquito-borne diseases (MBDs) \cite{dengue}. Factors like changing climate and ecological conditions, global travel and trade, human behavior, and rapid, unplanned urbanization significantly affect the seasonal and geographical spread of these vectors, impacting the transmission of their pathogens \cite{JRC119557, rs15245649}.

According to the European Center for Disease Prevention and Control (ECDC), the number of confirmed malaria cases reported in the European Union from 2008 to 2012 ranged between 5000 and 7000 \cite{europaMalaria5years}, whereas in 2018 it reached almost 8500 \cite{europaMalariaAnnual}. Europe faced a WNV epidemic in 2010 with 1016 cases \cite{europaWNV2010}, while in 2018 the cases rose to 1516 \cite{europaWNVAnnual}. Moreover, dengue cases increased nearly 450\% globally from 1990 to 2013 \cite{dengue_increase} with the total annual global cost of dengue illness estimated at 9 billion USD/year \cite{dengue_cost}.

In light of the surge in MBDs outbreaks, machine learning (ML) techniques and Earth Observation (EO) data are increasingly being utilized in the combat against MBDs. Several methods have been employed to tackle the problem. Following a straightforward approach, \cite{malaria_india, dengue_china} used Support Vector Machines to directly predict malaria and dengue human cases in India and China, respectively. Decision trees, which offer the advantage of prediction explainability, were employed to estimate the WNV risk across the USA on a county level \cite{WNV_USA}, while generalized additive models were used in \cite{WNV_USA2} to predict future WNV outbreaks in space and time in northern Great Plains of the USA.

Since the abundance of vector species plays a key role in disease outbreak emergence \cite{MA-MBDs}, a $k$-nearest neighbors regression prediction model was employed in Argentina to estimate the \textit{Aedes aegypti} oviposition activity \cite{MA-knn}, while a similar method using deep learning models for time series was adopted in Madeira Island, Portugal \cite{MA-dl}. In a more general approach, \cite{tsantalidou} deployed an area transferable and mosquito genus agnostic framework for predicting upcoming mosquito populations based on an XGBoost regression.

Nevertheless, the growing reliance on ML algorithms in critical sectors is accompanied by substantial risks. Since we deal with limited data, such as EO and entomological data in the field of public health-related applications, the need for accuracy and precision in these algorithms \cite{debie2021trust} becomes indispensable. Model reliability is particularly critical as the consequences of incorrect forecasts could have a significant impact on policy-making. A major concern is the possibility of these models to yield inaccurate predictions, especially when they encounter situations that diverge from their training data. Effective explanations of ML models applied to EO data, enhance usage and error identification, and are crucial when models depend on irrelevant factors \cite{debie2021trust, doshivelez2017rigorous, 10.1145/2939672.2939778}. This clarity is important, as users tend to distrust algorithms after erroneous decisions, often favoring human judgment despite a superior algorithmic performance \cite{10.1145/2939672.2939778, dietrvorst}.


In this paper, we propose a method to characterize the trustworthiness of predictions on health-related and EO data. Our method is applied to estimate the upcoming MA using EO tabular data for the areas of Veneto in Italy and Upper Rhine Valley in Germany, yielding promising results. Specifically, we leverage Latent Space (LS) representations generated by a Variational AutoEncoder to estimate the confidence of each MA prediction based on their distance. To the best of our knowledge, there is no extensive research in the field of similar VAE architectures related to emerging health-related applications, hence we propose a method that relies on the LS distance that can be employed in related applications.



\section{Data}
\label{sec:data}
The dataset incorporates EO data, which are sourced from different satellites, alongside entomological MA in situ information. The data of this research cover two areas of interest (AOI), namely the Veneto region in Italy and the Upper Rhine Valley in Germany.

\subsection{EO Data}\label{sec2.1}
EO data consists of environmental and topographic data for each of the AOIs.

\subsubsection{Environmental data}\label{sec2.1.1}
Land Surface Temperature (LST) measurements (in  $^\circ C$) from the MODIS sensors onboard TERRA and AQUA satellites are used. LST is estimated using the top-of-atmosphere brightness temperatures from the infrared bands of the satellite sensors. The product incorporated into the model is V6.0, which provides day and night LST measurements from satellite overpasses with a spatial resolution of 1 kilometer (km).
Sentinel 2 (10 m GSD, six-day revisit time) and Landsat TM 7 \& 8 are employed to calculate proxies for vegetation density, changes in vegetation water content, determination of vegetation water content, and mapping of built-up areas. The Integrated Multi-satellite Retrievals for GPM (IMERG) precipitation gridded dataset at a resolution of 0.1$^{\circ}$ $\times$ 0.1$^{\circ}$ is used to extract precipitation measurements.

\subsubsection{Topographic Data}\label{sec2.1.2}
The Digital Elevation Model (DEM) product that is used to generate parameters, such as elevation, slope, and aspect, is acquired from the Copernicus LMS with a spatial resolution of 25 m. For each trap site, the mean elevation, slope, and aspect are calculated within a buffer zone of 1 km around the point. WWFHydroSHEDS\footnote{https://www.hydrosheds.org/} and Copernicus land cover products are used to calculate hydrological features.

\subsection{Entomological Data}\label{sec2.2}
\begin{figure}[htb]
\begin{minipage}[b]{.48\linewidth}
  \centering
  \centerline{\epsfig{figure=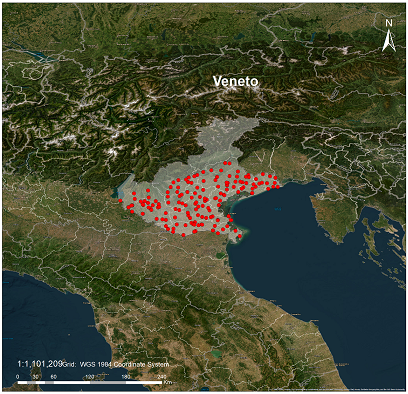,width=4cm}}
  \centerline{(a) Italy}\medskip
\end{minipage}
\hfill
\begin{minipage}[b]{.48\linewidth}
  \centering
  \centerline{\epsfig{figure=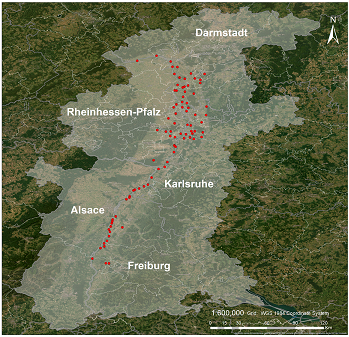,width=4cm}}
  \centerline{(b) Germany}\medskip
\end{minipage}
\caption{Entomological network in the two AOIs. Each dot represents the location of a mosquito trap.}
\label{fig:ent_network}
\end{figure}

In order to collect MA data, a systematic approach for entomological monitoring has been effective since 2010 in Europe, collecting data from stable station networks using CDC-CO\textsubscript{2} light traps and gravid traps roughly on an every other week basis, identifying the total number of mosquitoes and the number of mosquitoes that are tested positive to the pathogen. As an example, Figure \ref{fig:ent_network} depicts the entomological network in the two AOIs.

\section{Proposed Confidence Metric}
\label{sec:methods}
The objective of the paper is to define a method that characterizes the trustworthiness of a prediction before its evaluation. In other words, given the train set $X_{train}$ and the predictor $P(\cdot)$, we want to derive a confidence metric $\mathcal{C}$ that is informative regarding the prediction's expected error $\hat{e}$, for each unknown observation $X_{un}$. The confidence metric can be expressed mathematically as  

\begin{equation}
    \mathcal{C}(X_{un}|X_{train}, P(\cdot)) \propto \hat{e} ~ ~.
\end{equation}

Our approach capitalizes on Variational AutoEncoders (VAEs), where the encoder $\text{Enc}(\cdot)$ transforms the input observation $X$ (consisting of environmental, topological, and other information) to a new representation, $Z$, in the latent space, i.e., $\text{Enc}(X) = Z$. Then, the decoder $\text{Dec}(\cdot)$ uses the representation $Z$ as input to estimate the expected number of mosquitoes $D(Z) = \hat{y}$. Therefore, the predictor can be written as
\begin{equation}
    P(X) \equiv \text{Dec}(\text{Enc}(X)) = \hat{y} ~~.
\end{equation}

\begin{figure}[htbp]
\centering
\centerline{\epsfig{figure=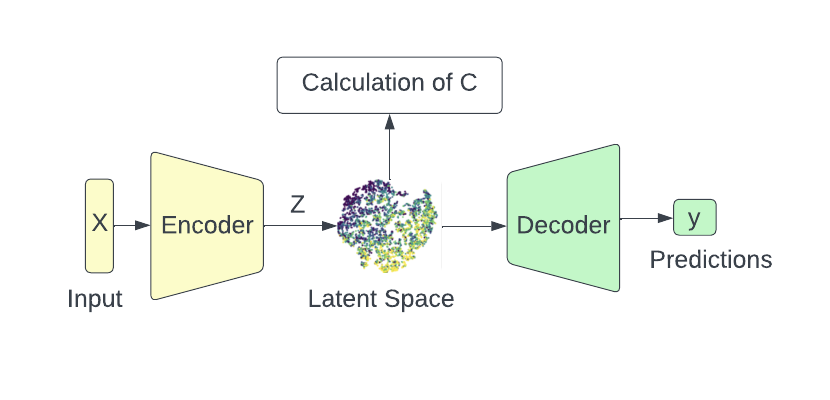, width=0.5\textwidth}}
\caption{The main components of our VAE architecture}.
\label{fig:meth}
\end{figure}

By taking advantage of the complex latent space representations, the key idea is to calculate $\mathcal{C}$ in the latent space of the VAE (see Figure \ref{fig:meth}).

\begin{figure}[htb]
\begin{minipage}[b]{.49\linewidth}
  \centering
  \centerline{\epsfig{figure=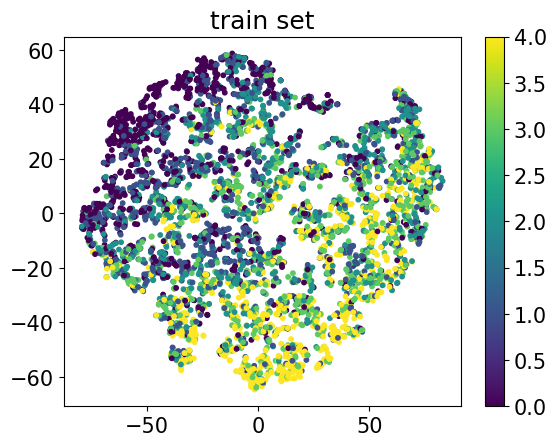,width=4cm}}
  \centerline{(a) Train Set}\medskip
\end{minipage}
\hfill
\begin{minipage}[b]{.49\linewidth}
  \centering
  \centerline{\epsfig{figure=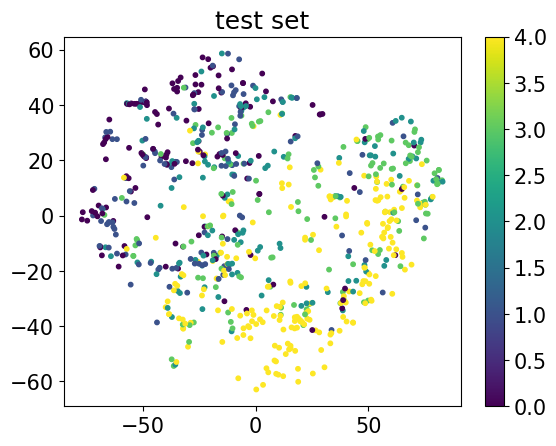,width=4cm}}
  \centerline{(b) Test Set}\medskip
\end{minipage}
\caption{T-SNE Visualization of the latent space for both train and test dataset for the Veneto Italy dataset.}
\label{fig:latent}
\end{figure}

Figure \ref{fig:latent} shows the projections in the latent space for both train and unknown (test) observations, once the predictor is trained with the available $X_{train}$. The color code depicts the prediction error of each observation. Additionally, in Figure \ref{fig:latent} we observe that the spatial distribution regarding the error follows the same trend in both train and test sets.

\subsection{Confidence metric $\mathcal{C}$ calculation}
Once the spatial distribution of the error is similar in both train and test representations in the latent space, we consider that the unknown observations that are closer to representations of training points $Z_{train}$ and do have prediction error less than a predefined threshold $T$, e.g., $T = <\hat{e}_{train}>$, are more trustworthy. We split the overall train representations according to their error as follows:
\begin{equation}
\!
\begin{aligned}
Z_{train}^{+} &= { Z_{train}: \lVert D(Z_{train}) - y\rVert \leq T } \\
Z_{train}^{-} &= { Z_{train}: \lVert D(Z_{train}) - y\rVert > T } ~~.
\end{aligned}
\end{equation}

Regarding the confidence distance metric $\mathcal{C}$, which should be informative regarding the prediction error, we calculate the distance of $j$-th unknown representation $z_{un}^j \in Z_{un}$ by taking the set of nearest $M$ representations that belongs in $Z_{train}^{+}$. Thus, $Z_j^+ = \{z_{j,1}^+, z_{j,2}^+, ..., z_{j,M}^+\} \in Z_{train}^{+}$.

So the distance $\mathcal{C}$ for the $j$-th unknown observation can be written as
\begin{equation}
    \mathcal{C}_j \equiv \frac{1}{M} \sum_{m=1}^{M} \left\lVert z_{un}^j - z_{j,m}^+ \right\rVert_2~~.
\end{equation}

\subsection{Evaluation Metrics}
To evaluate if the proposed confidence metric $\mathcal{C}$ is informative with the error, we measure its correlation
\begin{equation}
    r = corr(\mathcal{C},\hat{e}) ~~.
\end{equation}
Additionally, we measure the average error of the top $20\%$ of the unknown observations with the highest confidence and the average error of the bottom $20\%$ that we characterize as unreliable.

\begin{table*}[htbp]
\centering
\begin{tabular}{|l|ccc|ccc|}
\hline
                   & \multicolumn{3}{c|}{\textbf{Italy}}         & \multicolumn{3}{c|}{\textbf{Germany}}        \\ \hline
\textbf{Distance}   & \textbf{GS} & \textbf{FS} & \textbf{LS}     & \textbf{GS} & \textbf{FS} & \textbf{LS}     \\ \hline
\textbf{Overall MAE}   & 124.41      & 124.41      & 124.41 & 273.66      & 273.66    & 273.66 \\ \hline
\textbf{MAE most reliable  20\% of samples} & 120.96   & 104.10   & \textbf{59.33}  & 150.2   & 100.5   & \textbf{5.1}    \\ \hline
\textbf{MAE most unreliable  20\% of samples}  & 99.87    & 128.41   & \textbf{264.31} & 152.37   & 292.92  & \textbf{780.02} \\ \hline
\textbf{Correlation}  & -0.02       & 0.01        & \textbf{0.36}   & -0.05       & 0.09        & \textbf{0.46}   \\ \hline
\end{tabular}
\caption{Results in the dataset of Italy and Germany. The testing samples are sorted by distance in ascending order. MAE on the first 20\% of samples (low distance samples) and the last 20\% of samples (high distance samples) is calculated, as well as the correlation between the distance and the absolute error of the predicted MA.}
\label{tab:performance}
\end{table*}

\section{Results}
\label{sec:RES}
As mentioned above, the performance of our methodology is assessed on two AOI, namely Veneto in Italy and Upper Rhine Valley in Germany. In both AOIs, we train our models with historical data (2010-2020) and predict the upcoming MA population for the observations of 2021. The selection of the year 2021 as a testing period is not random and has to do with the fact that the summer of that year was predominantly marked by unusually high levels of rainfall and subsequent flooding in Germany, leading to extreme MA values. The exceptional weather conditions during that period adversely affected the model's predictive performance resulting in more uncertain estimations of MA.

Once the training process is over, we choose the threshold $T = <\hat{e}_{train}>$ and the number of nearest representations $M=3$ to calculate the $\mathcal{C}$ for each unknown observation.  

In order to compare the effectiveness of our confidence metric in the Latent Space (LS) in assessing the confidence of the predictions, besides LS distance, two more metrics are calculated in the Geographical Space (GS) and the Feature Space (FS). 

\textbf{Geographical Euclidean Distance, GS:} measures how close the test data points are to the training data in terms of geographical location. It calculates the Euclidean distances based on the geographical coordinates of the points. This metric is commonly used in practical EO applications.

\textbf{Feature Space Distance, FS:} measures how similar the test data points are to the training data in the original feature space. This involves calculating Euclidean distances again, but this time based on the feature vectors of the data points.

\begin{figure}[htb]
\begin{minipage}[b]{.48\linewidth}
  \centering
  \centerline{\epsfig{figure=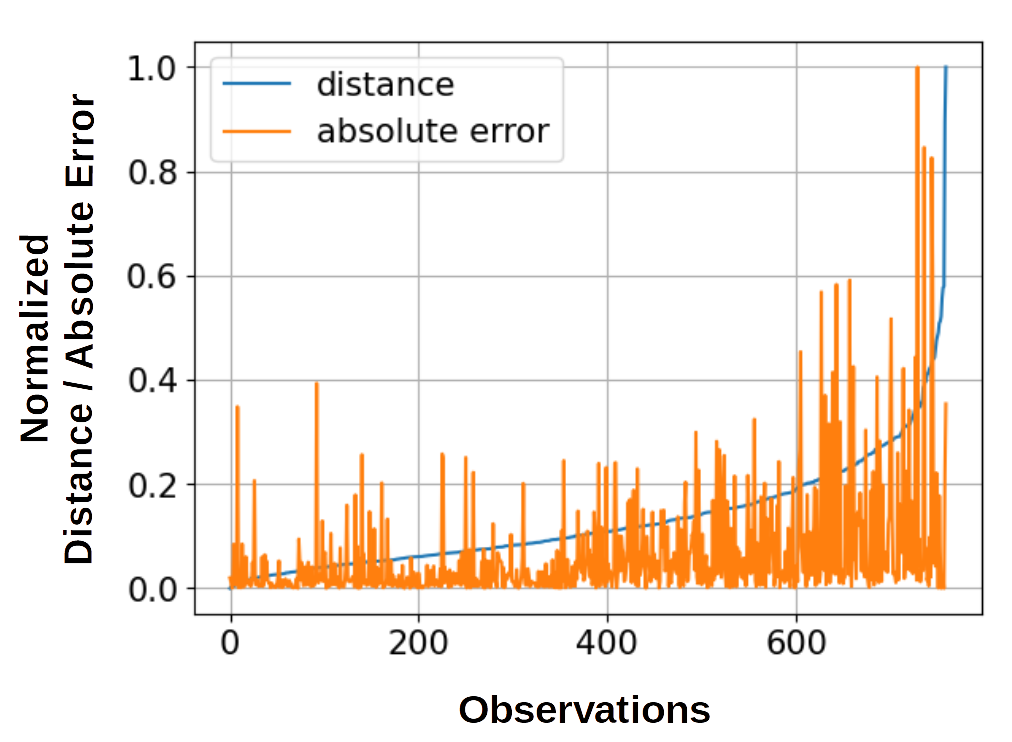,width=4cm}}
  \centerline{(a) Italy}\medskip
\end{minipage}
\hfill
\begin{minipage}[b]{.48\linewidth}
  \centering
  \centerline{\epsfig{figure=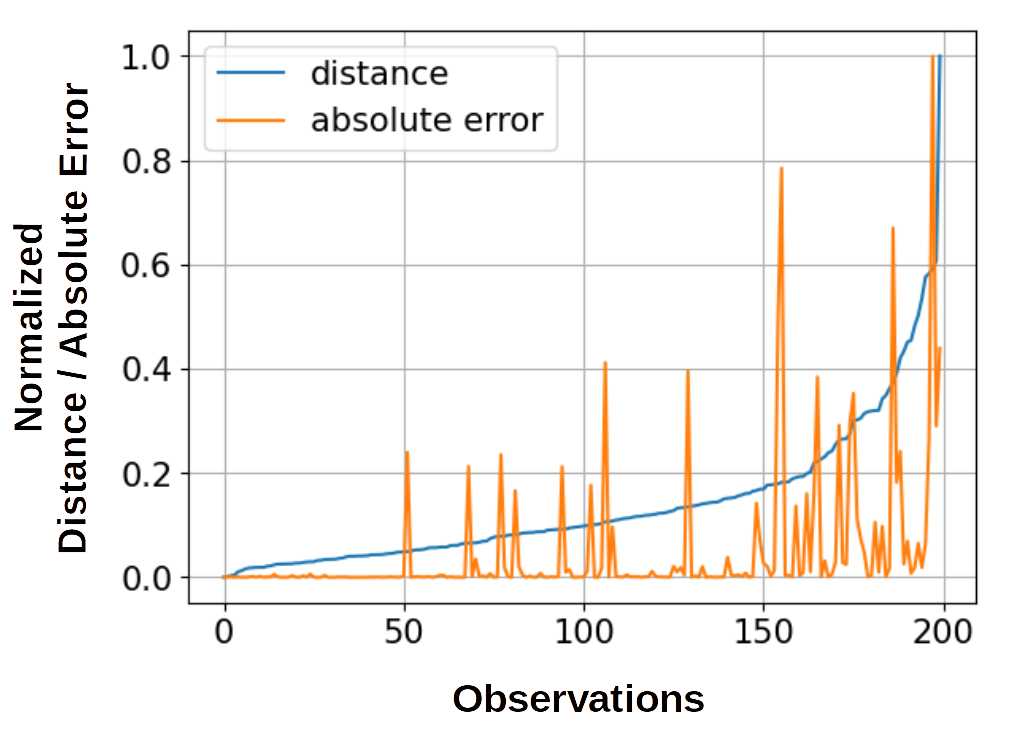,width=4cm}}
  \centerline{(b) Germany}\medskip
\end{minipage}
\caption{The corresponding absolute error and confidence $\mathcal{C}$ for the unknown observations.}
\label{fig:dist}
\end{figure}

Table \ref{tab:performance} compares the MAE and its correlation with the distance in the three different spaces for the datasets of Italy and Germany.

Ranking the predictions based on their GS distance led to poor separation between the two groups in terms of MAE, which is also indicated by the nearly zero correlation between the GS distance and the absolute error. Based on the aforementioned results, it can be concluded that the geographical location is irrelevant to the accuracy of the prediction. A similar behavior is indicated by the distance in the FS, as the correlation remains low. 

In sharp contrast, the correlation of the LS distance with absolute error is significantly higher, $0.36$ and $0.46$ in Italy and Germany, respectively, indicating that the higher the distance, the larger the error. Moreover, Figure \ref{fig:dist} illustrates the error for increasing distance in LS. An increasing trend of error is observed as the distance increases. Measuring the Euclidean distance in the constructed latent space enables us to rank the predictions based on their validity.



A higher correlation in LS implies that the distances in this space are more meaningful and representative of the actual relationships between data points. The model seems to be more 'confident' in its predictions when the LS distance is smaller, indicating a stronger and more reliable relationship between test and training sets in LS as compared to FS. The higher correlation in the LS suggests that the model is effectively capturing the underlying structure of the data. The latent space, being a lower-dimensional representation, likely distills the essential features of the data, leading to more meaningful and reliable distances.

\section{Discussion and Conclusions}
\label{sec:CON}
This work aimed to address the pressing need for confidence and trustworthiness in machine learning predictions, particularly within the scope of health-related and earth observation data. By designing and implementing a VAE model, we have demonstrated the capability of this model to distill complex data into a meaningful latent space. Our results distinctly highlight the correlation in latent space, which underpins the model's enhanced ability to capture the essential features of data and to provide a reliable confidence measure.

More importantly, our findings suggest that the distances within the latent space are more than mere numerical values; they are indicative of the underlying data structure and bear significant implications for prediction confidence. This is particularly evident when comparing the latent space correlation with that of the feature space, with the former showing higher values.

Our proposed methodology shows a new potential use of latent spaces, not only as a method for dimensionality reduction but also as a cornerstone for trustworthiness in AI/ML systems. It serves as a bridge between raw data and interpretable, trustworthy predictions, fostering user confidence and paving the way for the deployment of AI in sensitive domains.

Future work will aim to refine this approach, exploring the scalability of the method to other forms of data and regression tasks.

\section{ACKNOWLEDGEMENTS}
\label{sec:ACK}
I. Pitsiorlas and M. Kountouris have been supported by the SNS JU project ROBUST-6G under the EU’s Horizon programme Grant Agreement No. 101139068. A. Tsantalidou, G. Arvanitakis, and Ch. Kontoes thank the EYWA Project Consortium for its overall support.

\bibliographystyle{IEEEbib}
\bibliography{strings,refs}

\end{document}